# Particle Filtering for Attitude Estimation Using a Minimal Local-Error Representation: A Revisit

*Lubin Chang*

*Department of navigation engineering, Naval University of Engineering, China*

For spacecraft attitude representation, the quaternion is preferred over other representations due to its bilinear nature of the kinematics and the singularity-free property. However, when applied in the attitude estimation filters, its unity norm constraint can be easily destroyed by the quaternions averaging operation. Although the brute-force quaternion normalization can be implemented, this strategy is crude and may be not sufficient for some filters which necessitate the state covariance calculations [2-4]. In [1] a quaternion-based particle filtering (QPF) for spacecraft attitude estimation has been proposed based on the local/global attitude representation. For the local/global representation, the unconstrained three-dimensional representation, say modified Rodrigues parameter (MRP), is used for the local-error space while the quaternion is used for the global representation and propagation. With this ingenious structure, the unconstrained local-error representation is used directly as the state in the developed QPF, which can maintain the unity norm of the quaternion in a nature way. Therefore, there is no need to perform the brute-force quaternion normalization any more.

Since the MRP and quaternion based particles are used simultaneously in the filtering cycle of the QPF, they should be transformed to each other, which necessitates a fiducial attitude quaternion (see Eq. (18) in [1]). In the QPF, the estimated quaternion that is given by

$$\hat{\mathbf{q}}_{k+1}^{-} = \Omega\left[\hat{\boldsymbol{\omega}}_{k}^{+}\right]\hat{\mathbf{q}}_{k}^{+} \tag{1}$$

is used as the fiducial quaternion. The explicit definitions and expressions of the terms involved in Eq. (1) can be found in [1] and they are not presented for brevity. It is shown that

Eq. (1) inherits a similar manner as the extended Kalman filter (EKF) for state propagation [5]. It is well known that the particle filtering (PF) bears the superiority in terms of accuracy and robustness over other filtering algorithms with no exception of the EKF due to large amount of particles being used to make probabilistic inference. In this respect, the quaternion propagation in Eq. (1) has not made full use of the superiority of the PF. With this consideration, the estimated quaternion by the following quaternions averaging may be more preferred.

$$\tilde{\mathbf{q}}_{k+1}^{-} = \sum_{i=1}^{N} w_{k+1}^{(i)} \mathbf{q}_{k+1}^{(i)} \qquad (2)$$

where $\mathbf{q}_{k+1}^{(i)}$ are the quaternion based particles which can be obtained through Eq. (16) in [1] and $w_{k+1}^{(i)}$ are the corresponding weights. Unfortunately, as has been pointed out that the norm constraint of the quaternion estimate given by the averaging operation in Eq. (2) can be easily destroyed. Actually, this is just the motivation of the local/global representation structure being used, that is, avoiding the quaternions averaging operation is the filtering recursion.

If the norm constraint preserving quaternion can be derived using the quaternion based particles, it can be used to substitute the estimated quaternion in Eq. (1) as the fiducial attitude quaternion for the QPF. Fortunately, either the minimum mean-square error (MMSE) [6, 7] or maximum a posteriori (MAP) [6] approach can be used to derive a normalized quaternion estimate without any brute-force quaternion normalization. Since the MAP approach usually yields noisier estimates than the MMSE approach, the MMSE approach is more preferred. In this note, the QPF is revised and modified by making use of the MMSE approach to derive the fiducial attitude quaternion based on the aforementioned discussion.

When $N$ weighted quaternion based particles have been obtained through Eq. (16) in [1], the normalized quaternion estimate can be derived by solving the following maximization problem

$$\breve{\mathbf{q}}_{k+1}^{-} = \arg\max_{\mathbf{q}\in\mathbf{S}^3} \mathbf{q}^T M \, \breve{\mathbf{q}} \tag{3}$$

where $M$ is a $4\times 4$ matrix given by

$$M = \sum_{i=1}^{N} w_{k+1}^{(i)} \mathbf{q}_{k+1}^{(i)} \mathbf{q}_{k+1}^{(i)\,T} \tag{4}$$

Similarly to the well-known Davenport's q method, the average quaternion that solves the maximization problem of Eq. (3) is the normalized eigenvector of $M$ corresponding to the maximum eigenvalue.

With the normalized quaternion estimate $\breve{\mathbf{q}}_{k+1}^{-}$, Eq. (18) in [1] can be modified as

$$\delta\mathbf{q}_{k+1}^{(i)} \equiv \begin{bmatrix} \delta\boldsymbol{\rho}_{k+1}^{(i)} \\ \delta q_{4_{k+1}}^{(i)} \end{bmatrix} = \mathbf{q}_{k+1}^{(i)} \otimes \left(\breve{\mathbf{q}}_{k+1}^{-}\right)^{-1} \tag{5}$$

Meanwhile, the updated quaternion given by Eq. (23) in [1] should also be modified as

$$\hat{\mathbf{q}}_{k+1}^{+} = \delta\hat{\mathbf{q}}_{k+1}^{+} \otimes \breve{\mathbf{q}}_{k+1}^{-} \tag{6}$$

With the aforementioned modification, the QPF can not only maintain the virtue of preserving quaternion norm constraint naturally, but also make full use of the advantage of the PF in terms of accuracy and robustness.

Making use of the MMSE approach to derive a normalized quaternion estimate has been investigated in the PF in [6]. However, in [6] the quaternion is used directly as the state and the local/global representation structure has not been used. This is because that there is no need to calculate the state covariance in the designed PF in [6]. In contrast, the attitude-error covariance (for MRP based state) should be calculated to perturb the state particles in [1], which necessitates the unconstrained three-dimensional local-error representation. More specifically, the quaternion is a four-element parameterization that is used to represent the three dimensional attitude, the rank of the $4\times 4$ covariance matrix corresponding to the quaternion is virtually three. Therefore, carrying out square root decomposition or inversion on the quaternion covariance matrix, which are usually necessary procedures in filtering algorithms, can result in undesired numerical difficulty. In this respect, the $3\times 3$ full rank covariance matrix should be determined for the attitude (actually the attitude error) based state.

The concept of modification discussed in this paper has also been used to modify the UnScented Quaternion Estimator (USQUE) in [8]. Actually, the local/global representation structure is just popularized by USQUE in [9, 10] and has been extended to other nonlinear sampling based filtering algorithm with the QPF in [1] as a representation. Therefore, the concept of modification discussed in this note and [8] is promising to modify the local/global representation structure based filtering algorithm, especially the sampling based nonlinear Kalman filtering algorithms which necessitate the state covariance calculations [11-13].

## Acknowledgments

This work was supported in part by the National Natural Science Foundation of China (61304241, 61374206).

## References

[1] Cheng, Y., and Crassidis, J. L., "Particle Filtering for Attitude Estimation Using a Minimal Local-Error Representation," *Journal of Guidance, Control, and Dynamics*, Vol. 33, No. 4, July–Aug. 2010, pp. 1305–1310.

[2] Shuster, M. D., "A Survey of Attitude Representations," *Journal of the Astronautical Sciences*, Vol. 41, No. 4, Oct.–Dec. 1993, pp. 439–517.

[3] Markley, F. L., "Attitude Error Representations for Kalman Filtering," *Journal of Guidance, Control, and Dynamics*, Vol. 63, No. 2, 2003, pp. 311–317.

[4] Crassidis, J. L., Markley, F. L., and Cheng, Y., "A Survey of Nonlinear Attitude Estimation Methods," *Journal of Guidance, Control, and Dynamics*, Vol. 30, No. 1, Jan.–Feb. 2007, pp. 12–28.

[5] Crassidis, J. L., and Junkins, J. L., Optimal Estimation of Dynamic Systems, CRC Press, Boca Raton, FL, 2004, pp. 285–292.

[6] Oshman, Y., and Carmi, A., "Attitude Estimation from Vector Observations Using a Genetic-Algorithm-Embedded Quaternion Particle Filter," *Journal of Guidance, Control, and Dynamics*, Vol. 29, No. 4, July–Aug. 2006, pp. 879–891.

[7] Markley, F. L., Cheng, Y., Crassidis, J. L., and Oshman, Y., "Averaging Quaternions," *Journal of Guidance, Control, and Dynamics*, Vol. 30, No. 4, July–Aug. 2007, pp. 1193–1196.

[8] Chang, L. B., Hu, B. Q., and Chang, G. B., "Modified UnScented QUaternion Estimator based on Quaternion Averaging," *Journal of Guidance, Control, and Dynamics*, Vol.37, No. 1, Jan.–Feb. 2014, pp. 305-309.

[9] Crassidis, J. L., and Markley, F. L., "Unscented Filtering for Spacecraft Attitude Estimation," *Journal of Guidance, Control, and Dynamics*, Vol. 26, No. 4, July–Aug. 2003, pp. 536–542.

[10] Crassidis, J. L., "Sigma-Point Kalman Filtering for Integrated GPS and Inertial Navigation," *IEEE Transactions on Aerospace and Electronic Systems*, Vol. 42, No. 2, 2006, pp. 750–756.

[11] Jia, B., Xin, M., and Cheng, Y., "Sparse Gauss–Hermite Quadrature Filter with Application to Spacecraft Attitude Estimation," *Journal of Guidance, Control, and Dynamics*, Vol. 34, No. 2, March–April 2011, pp. 367–379.

[12] Jia, B., Xin, M., and Cheng, Y., "Anisotropic Sparse Gauss–Hermite Quadrature Filter," *Journal of Guidance, Control, and Dynamics*, Vol. 35, No. 3, May–June, 2012, pp. 1014–1023.

[13] Jia, B., and Xin, M., "Vision-Based Spacecraft Relative Navigation Using Sparse-Grid Quadrature Filter," *IEEE Transactions on Control Systems Technology*, Vol. 21, No. 5, 2013, pp. 1595-1606.